\documentclass[11pt,a4paper]{article}
\usepackage[hyperref]{acl2019}
\usepackage{times}
\usepackage{latexsym}

\usepackage{flushend}

\usepackage{url}

\usepackage{graphicx}
\usepackage{caption}
\usepackage{breakcites}
\usepackage{tabularx}
\usepackage{comment}
\usepackage{float} % prevent LaTeX from re-positioning tables. Use \begin{table}[H]
\restylefloat{table}
\aclfinalcopy % Uncomment this line for the final submission

\usepackage{amsthm,amsmath,amssymb,amsfonts,amscd}
\usepackage{graphicx}
\usepackage{enumerate}
\usepackage[all]{xy}
\usepackage{booktabs}
\usepackage[spanish,english]{babel}
\usepackage[utf8]{inputenc}
\usepackage[nottoc]{tocbibind}
\usepackage{multirow}
\usepackage{caption}
\usepackage{breakcites}
\usepackage{tabularx}
\usepackage{subcaption}
\title{Equalizing Gender Bias in Neural Machine Translation \\with Word Embeddings Techniques}

\author{Joel Escudé Font and Marta R. Costa-jussà\\
  Universitat Politècnica de Catalunya, 08034 Barcelona \\
  {\tt joel.escude@estudiant.upc.edu,marta.ruiz@upc.edu} \\}%\And
\date{}

\begin{document}
\maketitle
\begin{abstract}
Neural machine translation has significantly pushed forward the quality of the field. However, there are remaining big issues with the output translations and one of them is fairness. Neural models are trained on large text corpora which contain biases and stereotypes. As a consequence, models inherit these social biases. 
%An example of this is the fact that ``friend'' in the English sentence ``She works in a hospital, my friend is a nurse'' would be correctly translated to ``amiga'' (feminine of friend) in Spanish, while ``She works in a hospital, my friend is a doctor'' would be incorrectly translated to ``amigo'' (masculine of friend) in Spanish. We consider that this translation contains gender bias since it ignores the fact that, for both cases, ``friend'' is a female and translates by focusing on the occupational stereotypes, i.e. translating doctor as male and nurse as female.  %However, social constructs are learned by these models due to the fact that training data from human generated corpora contain biases such as gender stereotypes.
%%
Recent methods have shown results in reducing gender bias in other natural language processing tools such as word embeddings.
%, which is the task of representing words as vectors to extract similarities from them.
%%
%We take advantage of the fact that word embeddings are used in neural machine translation to propose one of the first attempts to debias neural machine translation. Specifically, we propose, experiment and analyze the integration of two debiasing techniques over GloVe embeddings in the Transformer translation architecture. 
We take advantage of the fact that word embeddings are used in neural machine translation to propose a method to equalize gender biases in neural machine translation using these representations.
We evaluate our proposed system on the WMT English-Spanish benchmark task, showing gains up to one BLEU point. As for the gender bias evaluation, we generate a test set of occupations and we show that our proposed system learns to equalize existing biases from the baseline system.
\end{abstract}

\section{Introduction}

Language is one of the most interesting and complex skills used in our daily life, and may even be taken for granted on our ability to communicate. However, the understanding of meanings between lines in natural languages is not straightforward for the logic rules of programming languages. 

Natural language processing (NLP) is a sub-field of artificial intelligence that focuses on making natural languages understandable to computers. 

Similarly, the translation between different natural languages is a task for Machine Translation (MT). Neural MT has shown significant improvements on performance using deep learning techniques, which are algorithms that learn abstractions from data. In recent years, these deep learning techniques have shown promising results in narrowing the gap between human-like performance with sequence-to-sequence learning approaches in a variety of tasks \cite{sutskever2014}, improvements in combination of approaches such as attention \cite{DBLP:journals/corr/BahdanauCB14} and translation systems algorithms like the Transformer \cite{vaswani2017}.

%% Bias
One downside of models trained with human generated corpora is that social biases and stereotypes from the data are learned \cite{madaan}. A systematic way of showing this bias is by means of word embeddings, a vector representation of words.
The presence of biases, such as gender bias, is studied for these representations and evaluated on crowd-sourced tests \cite{bolukbasi2016}.
The presence of biases in the data can directly impact downstream applications \cite{DBLP:conf/naacl/ZhaoWYOC18} and are at risk of being amplified \cite{DBLP:conf/emnlp/ZhaoWYOC17}.

% Another successful approach has been the popular Word Embeddings that have been used in many NLP applications as a low-dimensional semantic representation of words as vectors.

% {\color{red} falta context, passes de word embeddings a social bias?}

%However, social biases are learned from the corpora in which they are trained \cite{bolukbasi2016}. 

%% Objectives
The objective of this work is to study the presence of gender bias in MT and give insight on the impact of debiasing in such systems. An example of this gender bias is the word ``friend'' in the English sentence ``She works in a hospital, my friend is a nurse'' would be correctly translated to ``amiga'' (girl friend in Spanish) in Spanish, while ``She works in a hospital, my friend is a doctor'' would be incorrectly translated to ``amigo'' (boy friend in Spanish) in Spanish. We consider that this translation contains gender bias since it ignores the fact that, for both cases, ``friend'' is a female and translates by focusing on the occupational stereotypes, i.e. translating doctor as male and nurse as female. 

The main contribution of this study is providing progress on the recent detected problem which is gender bias in MT \cite{prates:2018}. The progress towards reducing gender bias in MT is made in two directions: first, we define a framework to experiment, detect and evaluate gender bias in MT for a particular task; second, we propose to use debiased word embeddings techniques in the MT system to reduce the detected bias. This is the first study in proposing debiasing techniques for MT.

%The following work is planned to be elaborated as a paper and submitted for review.

The rest the paper is organized as follows. Section \ref{sec:background} reports material relevant to the background of the study. Section \ref{sec:relwork} presents previous work on the bias problem. Section \ref{sec:methods} reports the methodology used for experimentation and section \ref{sec:expframe} details the experimental framework. The results and discussion are included in section \ref{sec:results} and section \ref{sec:conclusion} presents the main conclusions and ideas for further work.

\section{Background}
\label{sec:background}

This section presents the models used in this paper. First, we describe the Transformer model which is the state-of-the-art model in MT. Second, we report describe word embeddings and, then, the corresponding techniques to debias them. 

\subsection{Transformer}

%{\color{red} jo definiria primer encoder-decoder i després attention}

The Transformer \cite{vaswani2017} is a deep learning architecture based on self-attention, which has shown better performance over previous systems. It is more efficient in using computational resources and has higher training speed than previous recurrent \cite{sutskever2014,DBLP:journals/corr/BahdanauCB14} and convolutional models \cite{DBLP:journals/corr/GehringAGYD17}. %See Figure \ref{fig:transformer-arch} for a simplified version of its architecture.

%% Encoder-Decoder
%\vspace{8mm}
%\begin{figure}[H]
%    \centering
%    \includegraphics[width=0.5\textwidth]{visuals/imgs/encdec02.png}
%    \caption{Simplification of the Transformer architecture.}
%    \label{fig:transformer-arch}
%\end{figure}

The Transformer architecture consists of two main parts: an encoder and a decoder. The encoder reads an input sentence to generate a representation which is later used by a decoder to produce a sentence output word by word.

The input words are represented as vectors, word embeddings (more on this in section \ref{sec:we}) and then, positional embeddings keep track of the sequentiality of language.
%, to process language as a vector space representation which can have a fixed or variable length. Words surrounding another word determine its meaning and how it is represented in this space, thus context influences in deciding the appropriate meaning for a given task using such representation.
%General processing of sequential data is performed with recurrent neural netoworks (RNN) which perform several steps to provide a decision on distant words.
%The number of steps needed by a neural network is shown to influence negatively on the decision making process as the steps increase \cite{chapter-gradient-flow-2001:hochreiter}.
The Transformer architecture computes a reduced constant number of steps using a self-attention mechanism on each one. 
The attention score is computed for all words in a sentence when comparing the contribution of each word to the next representation.
New representations are generated in parallel for all words at each step .

Finally, the decoder uses self-attention in generated words and also uses the representations from the last words in the encoder to produce a single word each time.

\subsection{Word embeddings}
\label{sec:we}

% Definition
Word embeddings are vector representations of words. 
%This representation is less sparse and more expressive, opposite to discrete atomic symbols and one-hot vectors.
These representations are used in many NLP applications.
Based on the hypothesis that words appearing in same contexts share semantic meaning, this continuous vector space representation gathers semantically similar words, thus being more expressive than other discrete representations like one-hot vectors.

%% Properties: analogies and similarities
Arithmetic operations can be performed with these embeddings, in order to find analogies between pairs of nouns with the pattern ``A is to B what C is to D'' \cite{mikolov2013a}. For nouns, such as countries and their respective capitals or for the conjugations of verbs.
%See Figure \ref{fig:embedding-analogies}.

%% GloVe algorithms
%\subsection{GloVe}
While there are many techniques for extracting word embeddings, in this work we are using Global Vectors, or GloVe \cite{pennington2014}. Glove is an unsupervised method for learning word embeddings. This count-based method, uses statistical information of word occurrences from a given corpus to train a vector space for which each vector is related to a word and their values describes their semantic relations.

%% Encoder-Decoder
%\vspace{1mm}
%\begin{figure}[H]
%    \centering
%    \includegraphics[width=1.0\textwidth]{visuals/imgs/wemb.png}
%    \caption{Visualization of word embeddings. Source: \url{https://www.tensorflow.org/images/linear-relationships.png}.
%    \label{fig:embedding-analogies}}
%\end{figure}

%% Debiasing wemb
%%\subsection{Debiasing word embeddings}
\subsection{Equalizing biases in word embeddings}

The presence of biases in word embeddings is a topic of discussion about fairness in NLP.
%More specifically, gender stereotypes are learned from human generated corpora and a post-process method is proposed for debiasing previously trained word embeddings \cite{bolukbasi2016}.
More specifically, \citet{bolukbasi2016} proposes a post-process method for debiasing already trained word embeddings. 
%%and also quantifies gender stereotypes learned from human generated corpora.
%Work analyzing the impact of the bias in machine translation applications \cite{zhao2017}.
%Several debiasing approaches have been proposed.
%GN-GloVe is a method for generating gender neutral embeddings \cite{DBLP:conf/emnlp/ZhaoZLWC18}.
\cite{DBLP:conf/emnlp/ZhaoZLWC18} aims to restrict learning biases during the training of the embeddings to obtain a more neutral representation.
The main ideas behind these methods are described next.

%% Give perspective on how this will be used in the methodology 
%% Give details on how debiasing word embeddings works

% Detail of bolukbasi2016
%The first mentioned debiasing method  \cite{bolukbasi2016}, takes an existing word embedding model and process it to remove the gender stereotypes learned from the training corpus. 
%First, it identifies a gender subspace by using a list of gender-definitional word pairs. Second, it neutralizes the gender neutral words to zero in this subspace. Also, equalizes neutral words to be equidistant to all pair of words in an equality set.
%Equalization has the disadvantage of not preserving other properties independent of gender which may be needed for other implementations.

%\paragraph{Debiaswe}
\paragraph{Hard-debiased embeddings} 
\cite{bolukbasi2016} is a post-process method for debiasing word embeddings.
%\paragraph{\citet{bolukbasi2016}}
%It consists of two parts: 
First, the direction of the embeddings where the bias is present is identified. 
Second, the gender neutral words in this direction are neutralized to zero and also equalizes the sets  by making the neutral word equidistant to the remaining ones in the set.
The disadvantage of the first part of the process is that it can remove valuable information in the embeddings for semantic relations between words with several meanings that are not related to the bias being treated.

%% GN-GloVe algorithm

% Detail of zhao2018
%The other mentioned debiasing method \cite{zhao2018} is a modification of the GloVe embedding algorithm \cite{pennington2014}.
%It modifies the minimizing objective of the training to capture word proximity while restricting gender information to one dimension of the model to preserve neutral the remaining. It uses a set of male and female definitional words to define a metric.

\paragraph{GN-GloVe} \cite{DBLP:conf/emnlp/ZhaoZLWC18} is an algorithm for learning gender neutral word embeddings models. It is based on the GloVe representation \cite{pennington2014} and modified to learn such word representations while restricting specific attributes, such as gender information, to specific dimensions.
%%For a protected attribute like gender, the minimization objective is composed first by capturing the word proximity like GloVe \cite{pennington2014} and the second restricts gender information in a specific dimension so the other dimensions are neutral.
A set of seed male and female words are used to define metrics for computing the optimization and a set of gender neutral words is used for restricting neutral words in a gender direction.
%It aims to tackle with the limitations of Hard-Debiasing, the removal of valuable gender information when eliminating the bias, and the possible propagation of errors by the classifier which affects the performance of the model. 

\section{Related work}
\label{sec:relwork}

While there are many studies on the presence of biases in many NLP applications, studies of this type in MT are quite limited.

\citet{prates:2018} performs a case study on gender bias in machine translation. 
They build a test set consisting of a list of jobs and gender-specific sentences. 
Using English as a target language and a variety of gender neutral languages as a source, i.e. languages that do not explicitly give gender information about the subject, they test these sentences on the translating service Google Translate.
They find that occupations related to science, engineering and mathematics present a strong stereotype toward male subjects.
%However, late 2018, Google announced in their developers blog\footnote{\url{https://ai.googleblog.com/2018/12/providing-gender-specific-translations.html}} that efforts are put on providing gender-specific translations in Google Translate. 
%Thus, it gives now both the translation for female and male when translating from gender-neutral languages.

\citet{eva:2018} compile a large multilingual dataset on the politics domain that contains the speaker information. They specifically use this information to incorporate it in a MT system. Adding this information improves the translation quality.

Our contribution is different from previous approaches in the sense that we are explicitly proposing a gender-debiased approach for NMT as well as an specific analysis based on correference and stereotypes to evaluate the effectiveness of our technique.

\section{Methodology} % Methodology
\label{sec:methods}

% description of how to integrate pre-trained word embeddings in the transformer

%{\color{red} comences sense context...}

In this section, we describe the methodology used for this study. %First, we train different sets of words embeddings, using GloVe, GN-GloVe and using Hard-Debiasing on the previous GloVe embeddings.
%Second, we train different models with the Transformer using the set of word embeddings in the encoder and decoder, and only the encoder. The performance of the models is evaluated later in section \ref{sec:results} with the BLEU metric. %An analysis on the translations are compared in Chapter \ref{chapter:results} to study the impact of debiased embeddings on the models.
%The trained models are used so to study the gender bias in the output translations. The different approaches to the study are presented below.
The prior layer of both the encoder and decoder in the Transformer \cite{vaswani2017}, where the word embeddings are trained, is adapted to use pre-trained word embeddings. We train the system with different pre-trained word embeddings (based on GloVe \cite{pennington2014}) to have a set of models. The scenarios are the following:

\begin{itemize}
    \item No pre-trained word embeddings, i.e. they are learned within the training of the model.
    \item Pre-trained word embeddings learned from the same corpus. Specifically, GloVe, Hard-Debiased GloVe and Gender Neutral Glove (GN-GloVe) embeddings.
\end{itemize}

\begin{figure*}[h!]
\centering
\begin{minipage}{.33\textwidth}
  \centering
  \includegraphics[width=.85\linewidth]{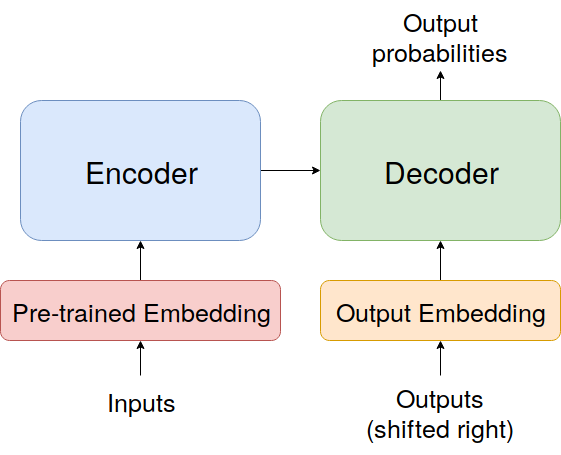}
\end{minipage}%
\begin{minipage}{.33\textwidth}
  \centering
  \includegraphics[width=0.85\linewidth]{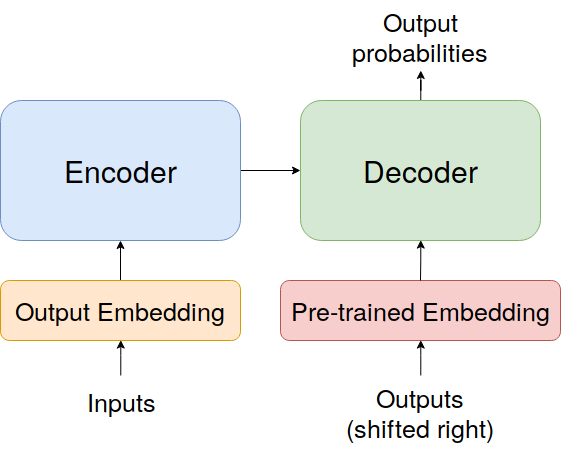}
\end{minipage}
\begin{minipage}{.33\textwidth}
  \centering
  \includegraphics[width=.90\linewidth]{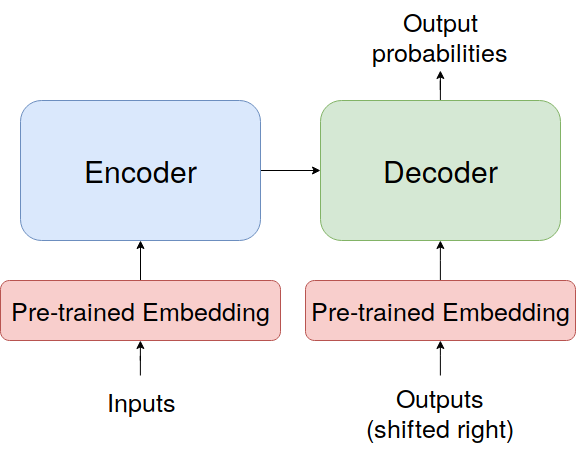}
\end{minipage}
  \caption{(Left) Pre-trained word embeddings in the encoder.
  (Center) Pre-trained word embeddings in the decoder.
  (Right) Pre-trained word embeddings in both the encoder and the decoder.}
  \label{fig:methods-encdec}
\end{figure*}

%The pre-trained embeddings are used for both the Encoder and Decoder sides, and also only for the Encoder side.
Also, the models with pre-trained embeddings given to the Transformer have three cases: using pre-trained embeddings only in the encoder side, see Figure \ref{fig:methods-encdec} (left), only in the decoder side, Figure \ref{fig:methods-encdec} (center), and both in the encoder and decoder sides, Figure \ref{fig:methods-encdec} (right).

%% Encoder-Decoder

%\item Both the Encoder and Decoder are given pre-trained embeddings. 
%\item Only the Encoder side is given pre-trained embeddings. 

%The implementation of the Transformer used for the experiments is OpenNMT \cite{klein2017}, an open-source toolkit for NMT which is actively maintained. 

\section{Experimental framework}
\label{sec:expframe}

% Summary
In this section, we present the experimental framework. We report details on the training of the word embeddings and the translation system. We describe the data related to the training corpus and test sets and the parameters. Also, we comment on the use of computational resources.

\subsection{Corpora}

The language pair used for the experiments is English-Spanish.
The training set consists of 16,554,790 sentences from a variety of sources including United Nations \cite{un}, Europarl \cite{europarl}, CommonCrawl and News available from the Workshop on Machine Translation (WMT) \footnote{\url{http://www.statmt.org/wmt13/}}.
The validation and test sets used are the \textit{newstest2012} (3,003 sentences) and \textit{newstest2013} (3,000 sentences), respectively, also from the same WMT workshop.
%For both the validation and test of the model, sets from the WMT translation of news task are used: newstest2012 and newstest2013, respectively. 
See Table \ref{tab:dataset} for the corpus statistics.

To study gender bias, we have developed an additional test set with custom sentences to evaluate the quality of the translation in the models. 
We built this test set using a sentence pattern ``\textit{I've known \{her, him, $<$proper noun$>$\} for a long time, my friend works as \{a, an\} $<$occupation$>$}.'' for a list of occupations from different professional areas.
We refer to this test as \textit{Occupations test}, their related sizes are also listed in Table \ref{tab:dataset} and sample sentences from this set are in Table \ref{tab:samples-occ}.
%{\color{red} falten les estadístiques d'aquest test! a la taula }
%See Table \ref{tab:samples-occ} also for sample sentences from this sets.
We use Spanish proper names to reduce ambiguity in this particular test.
These sentences are properly tokenized before using them in the test.

With these test sentences we see how ``friend'' is translated into its Spanish equivalent ``amiga'' or ``amigo'' which has a gender relation for each word, female and male, respectively. Note that we are formulating sentences with an ambiguous word ``friend'' that can be translated into any of the two words and we are adding context in the same sentence so that the system has enough information to translate them correctly. 
The list of occupations is from the U.S. Bureau of Labor Statistics\footnote{\url{https://www.bls.gov/cps/tables.htm\#empstat}}, which also includes statistical data for gender and race for most professions.
We use a pre-processed version of this list from \cite{prates:2018}.

\begin{table*}[h!]
\centering
\begin{tabular}{l}%[l]
%Sample sentences for the \textit{Occupations test} set\\
%\midrule
\hline
%%(En) I've known \{\textit{her, him, Mary, John}\} for a long time, my \textit{friend} works as an \textit{accounting clerk}. \\
%%(Es) \{\{\textit{La, Lo}\} conozco, Conozco a \{\textit{Mary, John}\}\} desde hace mucho tiempo, mi \{\textit{amiga, amigo}\} trabaja como \textit{contable}. \\
%%{\color{red} TODO simplify this table, even summarize to text (?)} \\
%%\hline
(En) I've known \textit{her} for a long time, my \textit{friend} works as an \textit{accounting clerk}. \\
(Es) \textit{La} conozco desde hace mucho tiempo, mi \textit{amiga} trabaja como \textit{contable}. \\
\hline
(En) I've known \textit{him} for a long time, my \textit{friend} works as an \textit{accounting clerk}. \\
(Es) \textit{Lo} conozco desde hace mucho tiempo, mi \textit{amigo} trabaja como \textit{contable}. \\
\hline
(En) I've known \textit{Mary} for a long time, my \textit{friend} works as an \textit{accounting clerk}. \\
(Es) Conozco a \textit{Mary} desde hace mucho tiempo, mi \textit{amiga} trabaja como \textit{contable}. \\
\hline
(En) I've known \textit{John} for a long time, my \textit{friend} works as an \textit{accounting clerk}. \\
(Es) Conozco a \textit{John} desde hace mucho tiempo, mi \textit{amigo} trabaja como \textit{contable}. \\
\hline
\end{tabular}
\caption{\label{tab:samples-occ} Sample sentences from the \textit{Occupations test} set. English (En) and Spanish (Es).}
\end{table*}

%{\color{red} falten les estadístiques d'aquest test! a la taula }

% Train: general.es-en.tc.clean.en,  general.es-en.tc.clean.es
% Validation newstest2012.tc.en, newstest2012.tc.es
% Test: newstest2013.tc.en,    newstest2013.tc.en
% WMT. Shared Task: Machine Translation of News. Translation task: NEWS

\begin{table*}[h!]
\centering
\begin{tabular}{llccccc}  
\toprule
Language & Data set & Num. of sentences & Num. of words & Vocab. size  \\
%%{\color{red} TODO summarize this table to text} \\
\midrule
\multirow{3}{*}{English (En) } 
%& Train & 16,554,790 & 427,614,914 & \\
%& Dev & 3,003 & 72,988 & \\
%& Test & 3,000 & 64,809 & \\
%& \multicolumn{2}{c}{Train} &
& Train &
16.6M & 427.6M & 1.32M \\
& Dev &
3k & 73k & 10k \\
%& \multirow{2}{*}{Test} &
& Test & 
3k & 65k & 9k \\
& \textit{Occupations test} & 
1k & 17k & 0.8k \\
\midrule
\multirow{3}{*}{Spanish (Es) } 
%& Train & 16,554,790 & 477,274,398 & \\
%& Dev & 3,003 & 78,887 & \\
%& Test & 3,000 & 70,540 & \\
& Train &
16.6M & 477.3M & 1.37M \\
& Dev &
3k & 79k & 12k \\
& Test & 
3k & 71k & 11k \\
& \textit{Occupations test} & 
1k & 17k & 0.8k \\
\bottomrule
\end{tabular}
\caption{English-Spanish data set.}
\label{tab:dataset}
\end{table*}

%% Models
\subsection{Models}

%% Embeddings
%\subsection{Word embeddings models}
%GloVe, GN-GloVe parameters. GloVe Accuracy.

%\subsubsection{Translation system}
% Parameters Transformer OpenNMT WMT
The architecture to train the models for the translation task is the Transformer \cite{vaswani2017} and we used the implementation provided by the OpenNMT toolkit\footnote{\url{http://opennmt.net/}}. The parameter values used in the Transformer are the same as proposed in the OpenNMT baseline system. Our baseline system is the Transformer witout pre-trained word embeddings.

 %and listed in Table \ref{tab:param-transf}.
Additionally, OpenNMT has built-in tools for training with pre-trained embeddings. 
In our case, these pre-trained embeddings have been implemented with the corresponding github repositories in GloVe\footnote{\url{https://github.com/stanfordnlp/GloVe}}, Hard-Debiasing with Debiaswe\footnote{\url{https://github.com/tolga-b/debiaswe}} and GN-GloVe\footnote{\url{https://github.com/uclanlp/gn_glove}}.
% Pre-trained embeddings
% {\color{red} aquest parraf no l'entenc aquí... no ha d'anar amb els embeddings?}
%The debiasing of the word embeddings uses files with definitional pairs of words to \cite{bolukbasi2016}.
%The pre-trained embeddings used in the model \cite{qi2018}. 

% http://opennmt.net/OpenNMT-py/FAQ.html#how-do-i-use-the-transformer-model-do-you-support-multi-gpu
\begin{comment}
\begin{table*}[h!]
\centering
\begin{tabular}{lc|lc}
\toprule
Parameter & Value & Parameter & Value \\
\midrule
Layers & 6 &
%RNN size & 512 \\
Word vec. size & 512\\ 
Transformer ff & 2048 \\
Heads & 8 &
Encoder type & transformer \\
Decoder type & transformer & 
Position encoding & (used) \\
Train steps & 200000 &
Max. generator batches & 2 \\
Dropout & 0.1 &
Batch size & 4096 \\
Batch type & tokens &
Normalization & tokens \\
Accum. count & 2 &
Optim. & adam \\
Adam beta2 & 0.998 &
Decay method & noam \\
Warmup steps & 8000 &
Learning rate & 2 \\
Max. grad. norm. & 0 &
Param. init. & 0 \\
Param. init. glorot & (used) & 
Label smoothing & 0.1 \\
Valid. steps & 10000 &
GPUs & 4 \\
\bottomrule
\end{tabular}
\caption{Parameter values used in the Transformer.}
\label{tab:param-transf}
\end{table*}
\end{comment}

%\subsubsection{Word Embeddings: Glove, Hard-Debiased GloVe and GN-GloVe}
% Parameters GloVe, GN-GloVe
The GloVe and GN-GloVe embeddings are trained from the same corpus presented in the previous section.
%%using GloVe \cite{pennington2014} and GN-GloVe \cite{DBLP:conf/emnlp/ZhaoZLWC18}. 
We refer to the method from \citet{bolukbasi2016} applied to the previously mentioned GloVe embeddings as Hard-Debiased GloVe.
The dimension of the vectors is settled to 512 (as standard) and kept through all the experiments in this study. The parameter values for training the word embedding models are shown in Table \ref{tab:param-wemb}. 

\begin{table}[h!]
\centering
\begin{tabular}{lcc}  
\toprule
Parameter & Value &  \\
\midrule
Vector size & 512 \\
Memory & 4.0 \\
Vocab. min. count & 5 \\
Max. iter. & 15 \\
Window size & 15 \\
Num. threads & 8 \\
X max. & 10 \\
Binary & 2 \\
Verbose & 2 \\
\bottomrule
\end{tabular}
\caption{Word Embeddings Parameters.}
\label{tab:param-wemb}
\end{table}

%{\color{red} falten els parametres de Hard-Debiased GloVe}

\citet{bolukbasi2016} uses a set of words to define the gender direction and to neutralize and equalize the bias from the word vectors.
Three set of words are used:
One set of ten pairs of words such as \textit{woman-man, girl-boy, she-he} are used to define the gender direction.
Another set of 218 gender-specific words such as \textit{aunt, uncle, wife, husband} are used for learning a larger set of gender-specific words.
Finally, a set of crowd-sourced male-female equalization pairs such as \textit{dad-mom, boy-girl, granpa-grandma} that represent gender direction are equalized in the algorithm.
In fact, for the English side, the gendered pairs used are the same as identified in the crowd-sourcing test by \citet{bolukbasi2016}.
%%Hard-Debiased GloVe and GN-GloVe files...
For the Spanish side, the sets are translated manually and modified when necessary to avoid non-applicable pairs or unnecessary repetitions. 
The sets from \citet{DBLP:conf/emnlp/ZhaoZLWC18} are similarly adapted to the Spanish language.

%% Language model
%\subsection{Transformer models}

To evaluate the performance of the models we use the BLEU metric \cite{papineni2002}. This metric gives a score for a predicted translation set compared to its expected output.

\subsection{Hardware resources}

The GPUs used for training are separate groups of four NVIDIA TITAN Xp and NVIDIA GeForce GTX TITAN. The duration time for training is approximately 3 and 5 days, respectively. In the implementation, the model is set to accumulate the gradient two times before updating the parameters, which simulates 4 more GPUs during training giving a total of 8 GPUs.%, however the training takes longer.

\section{Results}
\label{sec:results}

%\section{Evaluation}
In this section we report results on translation quality and present an analysis on gender bias.

\subsection{Translation} 
 
For the test set \textit{newstest2013}, BLUE scores are given in Table \ref{tab:bleu-newstest2013}. 
Pre-trained embeddings are used for training in three scenarios: in the encoder side (Enc.), in the decoder side (Dec.) and in both the encoder and decoder sides (Enc./Dec.). 
These pre-trained embeddings are updated during training. 
We are comparing several pre-trained embeddings against a baseline system (`Baseline' in Table \ref{tab:bleu-newstest2013}) which does not include pre-trained embeddings (neither on the encoder nor the decoder).

%For our study, we report results on pre-trained embeddings updated during training. 

%To evaluate the presence of gender bias on the system, we perform a qualitative analysis with a custom test set.

%The models with these fixed pre-trained debiased embeddings tend to show a slight decrease in performance and are not further evaluated for the gender bias experiments. 
%There is only one exception where fixed embeddings performed better than updated and it is for a non-debiased method. 
%They are left in the table for comparison. 
%The models trained with updated pre-trained embeddings used for a qualitative analysis on the impact of using debiased word embeddings in the translation system. 

%% BLEU
\begin{table}[h!]
\centering
\begin{tabular}{lccc}
    \toprule
    Baseline & & & 29.78 \\
    \midrule
    Pre-trained emb. & Enc. & Dec. & Enc./Dec. \\
    %& Enc & Dec & Enc+Dec \\
    \midrule
    GloVe & 30.21 & 30.24 & 30.62 \\
    %GloVe Hard-Debiased & 30.16 & 30.09 & 29.95 \\
    GloVe Hard-Deb. & 30.16 & 30.09 & 29.95 \\
    GN-GloVe & 29.12 & 30.13 & \textbf{30.74} \\
    \bottomrule
\end{tabular}
\caption{BLEU scores for the \textit{newstest2013} test set. English-Spanish. Pre-trained embeddings are updated during training. In bold best results.}
\label{tab:bleu-newstest2013}
\end{table}

For the studied cases, values do not differ much.
Using pre-trained embeddings can improve the translation, which is coherent with previous studies \cite{qi-EtAl:2018:N18-2}. Furthermore, debiasing with GN-GloVe embeddings keeps this improvement and even increases it when used in both the encoder and  decoder sides. 
%Note that best results are for the case of GN-Glove (in the encoder and the decoder) that we achieve 1 BLEU point improvement, 
We want to underline that these models do not decrease the quality of translation in terms of BLEU when tested in a standard MT task.
Next, we show how each of the models performs on a gender debiasing task.

%The statistical significance test with the bootstrap method \cite{DBLP:conf/emnlp/koehn2004} is used to compare BLEU scores between different models with same test sets. P-values lower than 0.1 are marked with *.
%The computes p-values are given in Table \ref{tab:bootstrap-newstest2013}.
%{\color{red} on estan els results d'aixo?}

%% p-values table
%\input{visuals/table-pvalues.tex}

%\section{Analysis}
%\section{Discussion}

%% Occupation tests

%Using a list of occupations from the US Bureau \footnote{\url{https://www.bls.gov/cps/tables.htm\#empstat}} we built a test set with a sentence patter ``\textit{I've known \{her, him, $<$proper noun$>$\} for a long time, my friend is \{a, an\} $<$occupation$>$.}''
%With these test sentences we see how ``friend'' is translated into its Spanish  equivalent ``amiga'' or ``amigo'' which has a gender relation for each word, female and male, respectively. 

\subsection{Gender Bias}

A qualitative analysis is performed on the \textit{Occupations test} set. Examples of this test set are given in Table \ref{tab:samples-occ}. %{\color{red} ADD such a table, with examples of translations}. 
The sentences of this test set contain context information for predicting the gender of the neutral word ``friend'' in English, either ``amigo'' or ``amiga'' in Spanish. 
The lower the bias in the system, the better the system will be able to translate the gender correctly. 
See Table \ref{tab:her-him} for the percentages of how ``friend'' is predicted for each model. 
%Note that we are using ``updated'' embeddings, since they are the best quality systems from Table \ref{tab:bleu-newstest2013}.

``Him'' is predicted at almost 100\% accuracy for all models. However not all occupations are well translated. 
On the other hand, the accuracy drops when predicting the word ``her'' on all models.
%the gender the accuracy of this task is not as precise as its counterpart and it shows a slight decrease in accuracy for all models.
When using names, the accuracy is even lower for ``Mary'' opposite to ``John''. 
%It is worth mentioning that proper names can also introduce racial stereotypes as shown by \citet{bolukbasi2016}.

%For the case using a person names (instead of the pronoun as coreferent with ``friend''), the accuracy of this task shows further biases. However, it is worth mentioning that proper names can induce another kind of bias such as racial stereotypes.

%% Occupations
\begin{table*}[h!]
\centering
\begin{tabular}{lrrrrrr}  
\toprule
\multirow{2}{*}{Pre-trained embeddings} & 
\multicolumn{1}{c}{her} &
\multicolumn{1}{c}{him} &
\multicolumn{1}{c}{Mary} &
\multicolumn{1}{c}{John} \\
& amiga  &  amigo 
& amiga  & amigo \\
\midrule
None & 99.8  & 99.9 & 69.5  & 99.9 \\ 
GloVe (Enc.) & 2.6  & 100.0 & 0.0 & 100.0 \\
GloVe (Dec.) & 95.0  & 100.0 & 4.0 & 100.0 \\
GloVe (Enc./Dec.)& \textbf{100.0} & 100.0 & 90.0 & 100.0 \\ 
GloVe Hard-Debiased (Enc.) & \textbf{100.0} & 100.0 & 99.5 & 100.0 \\ 
GloVe Hard-Debiased (Dec.) & 12.0  & 100.0 & 0.0 & 100.0 \\
GloVe Hard-Debiased (Enc./Dec.) & 99.9 & 	100.0 & \textbf{100.0}  & 99.9 \\ 
GN-GloVe (Enc.) & \textbf{100.0} & 100.0 & 7.7 & 100.0 \\ 
GN-GloVe (Dec.) & 97.2 & 100.0 & 51.8 & 100.0 \\
GN-GloVe (Enc./Dec.) & 99.6  & 100.0 & 56.4 & 100.0 \\ 
\bottomrule
\end{tabular}
\caption{Percentage of ``friend'' being translated as ``amiga'' or ``amigo'' in test sentences with female-male pronouns and proper names for the \textit{Occupations test}. 
%The pre-trained embeddings are updated during training.
Best results in bold.}
\label{tab:her-him} 
\end{table*}

Note that gender debiasing is shown by augmenting the percentage of ``amiga'' in the translation in the presence of the female pronoun while keeping the quality of translation (coherently with generic results in Table \ref{tab:bleu-newstest2013}). 
Based on accuracy values from Table \ref{tab:her-him}, the most neutral system is achieved with GloVe and also with Hard-Debiased GloVe pre-trained embeddings. The accuracy improves by 30 percentage points compared to the baseline system and over 10 percentage points compared to the non-debiased pre-trained word embeddings.

%{\color{red} add table with examples as in the master thesis}
%{\color{red} Two tables added. Comments on the examples are pending.}

The quality of the translation also depends on the professions from the \textit{Occupations test} and its predicted gender.
%The quality of the translations not only depends on the gender of the subject but also on the prediction of the profession's gender in our \textit{Occupations test}. 
%Since for almost all professions the model has no problem when predicting the gender for the pronoun ``him'' in Spanish, we focus the following analysis on the pronoun ``her''. 
Again, the system has no problem predicting the gender of professions in the context of ``him'', so we focus the analysis on the context of ``her''.
With GN-GloVe pre-trained embeddings both in the encoder and decoder sides, the model shows a higher accuracy when predicting the gender of a profession in Spanish. 
Specifically, for technical professions such as ``criminal investigator'', ``heating mechanic'', ``refrigeration mechanic'' and others such as ``mine shuttle car operator''.
See Table \ref{tab:pred-operator} for the prediction on this last profession.

\begin{table*}[h!]
\centering
\begin{tabular}{ll}
\toprule
Pre-trained word embeddings & 
Prediction
\\
%%{\color{red} TODO simplify, summarize to text}
&
\textit{\textbf{La} conozco desde hace mucho tiempo,}
\\
\midrule
None & \textit{mi amigo trabaja como mecánico de refrigeración.}
\\
GloVe (Enc) & \textit{mi \textbf{amiga} trabaja como mecánico de refrigeración.} \\
GloVe (Dec) & \textit{mi \textbf{amiga} trabaja como mecánico de refrigeración.} \\
GloVe (Enc+Dec) & \textit{mi \textbf{amiga} trabaja como mecánico de refrigeración.} \\
GloVe Hard-Debiased (Enc) & \textit{mi amigo trabaja como mecánico de refrigeración.} \\
GloVe Hard-Debiased (Dec) & \textit{mi \textbf{amiga} trabaja como mecánico de refrigeración.} \\
GloVe Hard-Debiased (Enc+Dec) & \textit{mi \textbf{amiga} trabaja como mecánico de refrigeración.} \\
GN-GloVe (Enc) & \textit{mi \textbf{amiga} trabaja como mecánico de refrigeración.} \\
GN-GloVe (Dec) & \textit{mi \textbf{amiga} trabaja como mecánico de refrigeración.} \\
GN-GloVe (Enc+Dec) & \textit{mi \textbf{amiga} trabaja como \textbf{mecánica de refrigeración}.} \\
\midrule
Reference & \textit{mi amiga trabaja como mecánica de refrigeración.} \\
\bottomrule
\toprule
Pre-trained word embeddings & 
Prediction \\
%%{\color{red} TODO simplify, summarize to text}
& \textit{\textbf{La} conozco desde hace mucho tiempo,} \\
\midrule
None & \textit{mi \textbf{amiga} trabaja como operador de un coche de enlace a las minas.} \\
GloVe (Enc) & \textit{mi amigo trabaja como operador del transbordador espacial.} \\
GloVe (Dec) & \textit{mi \textbf{amiga} trabaja como un operador de transporte de camiones.} \\
GloVe (Enc+Dec) & \textit{mi \textbf{amiga} trabaja como un operator de coches.} \\
GloVe Hard-Debiased (Enc) & \textit{mi \textbf{amiga} trabaja como mine de minas.} \\
GloVe Hard-Debiased (Dec) & \textit{mi amigo trabaja como un operador de transporte de coches para} \\
& \textit{las minas.} \\
GloVe Hard-Debiased (Enc+Dec) & \textit{mi \textbf{amiga} trabaja como un operator de coches.} \\
GN-GloVe (Enc) & \textit{mi \textbf{amiga} trabaja como operador de ómnibus de minas.} \\
GN-GloVe (Dec) & \textit{mi \textbf{amiga} trabaja como un operador de transporte para las minas.} \\
GN-GloVe (Enc+Dec) & \textit{mi \textbf{amiga} trabaja como \textbf{operadora de transporte de minas}.} \\
\midrule
Reference & \textit{mi amiga trabaja como operadora de vagones de minas.} \\
\bottomrule
\end{tabular}
\caption{\label{tab:pred-operator} 
Spanish predictions for the test sentences ``\textit{I've known her for a long time, my friend works as a refrigeration mechanic.}''
``\textit{I've known her for a long time, my friend works as a mine shuttle car operator.}''.
Best results in bold.
}
\end{table*}

\section{Conclusions and further work}
\label{sec:conclusion}

%% Conclusion
%Global and specific conclusions on the topic and with relation to the objective in the introduction.

Biases learned from human generated corpora is a topic that has gained relevance over the years. Specifically, for MT, studies quantifying gender bias present in news corpora and proposing debiasing approaches for word embedding models have shown improvements on this matter.

We studied the impact of gender debiasing on neural MT. We trained sets of word embeddings with the standard GloVe algorithm. Then, we debiased the embeddings using a post-process method \cite{bolukbasi2016} and also trained a gender neutral version \cite{DBLP:conf/emnlp/ZhaoZLWC18}. We used all these different models on the Transformer \cite{vaswani2017}. Experiments were reported on using these embeddings on both the encoder and decoder sides, or only the encoder or the decoder sides. 

The models were evaluated using the BLEU metric on the standard task of the WMT \textit{newstest2013} test set. BLEU performance increase when using pre-trained word embeddings and it is slightly better for the debiased models.

%To study the bias for the translations of the models, we developed a specific test set. 
In order to study the bias on the translations, we evaluate the systems on a custom test set composed of occupations.
This set consists of sentences that include context of the gender of the ambiguous ``friend'' in the English-Spanish translation.  This word can be translated to feminine or masculine and the proper translation has to be derived from context. We verified our hypothesis that consisted on the fact that if the translation system is gender biased, the context is disregarded, while if the system is neutral, the translation is correct (since it has the information of gender in the sentence). Results show that the male pronoun is always identified, despite not all occupations are well translated, while the female pronoun has different ratio of appearance for different models. 
In fact, the accuracy when predicting the gender for this test set is improved for some settings, when using the debiased and gender neutral word embeddings.
%In fact, we achieve 100\% accuracy of translation on these pronouns, both on the encoder and decoder sides, when using the debiased word embeddings. 
Also, as mentioned, this system slightly improves the BLEU performance from the baseline translation system. 
Therefore, we are ``equalizing'' the translation, while keeping its quality. 
Experimental material from this paper is available online \footnote{https://github.com/joelescudefont/genbiasmt}.
%It is worth mentioning that biases from the training data are possible still learned even using pre-trained embeddings.
%More research on the topic of debiasing is needed in NLP and Machine Translation. 

%\vspace{5mm}
%% Future work
%\section{Future work}

% Further work on tackling biases is a topic in NLP applications and Machine Translation that has been gaining importance over the last few years.
% Erasing biases is crucial for MT models to keep providing a positive impact on society by preventing, either harmful or neglected models, to shred the social fabric.

%As for future work in the topic of debiasing in Machine Translation and fairness in Machine Learning,

As far as we are concerned, this is one of the pioneer works on proposing gender debiased translation systems with word embedding techniques.
% Future work on the topic of debiasing translation algorithms is required.  %The kind of bias being studied for debiasing is one proposed direction.  Another direction is the type of corpora used for training the models.
%%

We did our study in the domain of news articles and professions. However, human corpora has a broad spectrum of categories, as an instance: industrial, medical, legal that may rise other biases particular to each area. Also, other language pairs with different degree in specifying gender information in their written or spoken communication could be studied for the evaluation of debiasing in MT. 
Furthermore, while we studied gender as a bias in MT, other social constructs and stereotypes may be present in corpora, whether individually or combined, such as race, religious beliefs or age; this being just a small subset of possible biases which will present new challenges for fairness both in machine learning and MT.

\section*{Acknowledgments}

This work is supported in part by the Spanish Ministerio de Econom\'ia y Competitividad, the European Regional  Development  Fund  and  the  Agencia  Estatal  de  Investigaci\'on,  through  the  postdoctoral  senior grant Ram\'on y Cajal, the contract TEC2015-69266-P (MINECO/FEDER,EU) and the contract PCIN-2017-079 (AEI/MINECO).

%The acknowledgments should go immediately before the references.  Do not number the acknowledgments section ({\em i.e.}, use \verb|\section*| instead of \verb|\section|). Do not include this section when submitting your paper for review.

%\bibliography{acl2018}
\bibliography{acl2019}
\bibliographystyle{acl_natbib}

\end{document}